\documentclass[10pt,twocolumn,letterpaper]{article}

\usepackage{wacv}              

\usepackage{graphicx}
\usepackage{subcaption}
\usepackage{multirow}

\definecolor{wacvblue}{rgb}{0.21,0.49,0.74}
\usepackage[pagebackref,breaklinks,colorlinks,allcolors=wacvblue]{hyperref}

\def\wacvPaperID{*****} 
\def\confName{WACV}
\def\confYear{2027}

\title{Distribution Shift Detection in Remote Sensing Imagery using Machine Learning Models}

\author{Abdullah Almansour\\
Carnegie Mellon University\\
Pittsburgh, USA\\
{\tt\small aalmanso@andrew.cmu.edu}
\and
Ozan K. Tonguz\\
Carnegie Mellon University\\
Pittsburgh, USA\\
{\tt\small tonguz@andrew.cmu.edu}
}

\begin{document}
\maketitle

\begingroup
  \renewcommand{\thefootnote}{}
  \footnotetext[0]{Accepted as a short paper at the 4th Workshop on Computer Vision for Earth Observation (CV4EO), held in conjunction with WACV 2027.}
\endgroup

\begin{abstract}
Remote sensing imagery acquired by Unmanned Aerial Vehicles (UAVs) supports wide range of applications; however, deploying machine learning (ML) models on resource-constrained aerial platforms introduces challenges beyond predictive accuracy, particularly when deployment data differ from the distributions observed during training. Such shifts are common in remote sensing due to geographic, sensor, seasonal, and other visual variations. A practical solution is to equip deployed models with mechanisms that detect emerging distribution shifts and support subsequent responsive actions. This paper presents an empirical case study of three open-source detection tools: (i) Alibi-Detect; (ii) NannyML; and (iii) PyTorch-OOD. Using cross-dataset remote-sensing shifts, we evaluate both detection effectiveness and computational latency of these tools. Our results reveal differences in detection behavior and processing cost across methods and implementations, highlighting trade-offs between detection capability and response-time. These findings motivate further work toward lightweight, and latency-aware monitoring mechanisms for resource-constrained systems.
\end{abstract}
    
\section{Introduction}
\label{sec:intro}

Over the past decades, machine learning (ML) models have achieved remarkable progress across a wide range of applications \cite{ref1, ref2}. However, strong performance on evaluation benchmarks does not necessarily translate to reliable operation after deployment. One source of model performance degradation is distribution shift, which occurs when the data encountered after deployment follow a different distribution from the data used during model training \cite{ref3}. Although numerous approaches have been proposed to improve models' generalization \cite{ref4, ref5}, the problem remains as a challenge in practical deployment. A potential solution is to monitor a model after deployment and detect the emergence of distribution shift \cite{tonguz2025using, taschin2025distribution, tonguz2025measurement}. Although several frameworks provide such capabilities \cite{ref6, ref7, ref8, ref9, ref99, ref10}, evaluating these tools only according to their detection accuracy overlooks an important aspect, which is the time required to obtain the detection decision. This becomes particularly relevant when monitoring is performed on resource-constrained systems, where computation may need to occur onboard under limited resources and strict response-time.

\begin{figure}[t]
  \centering
   \includegraphics[width=0.8\linewidth]{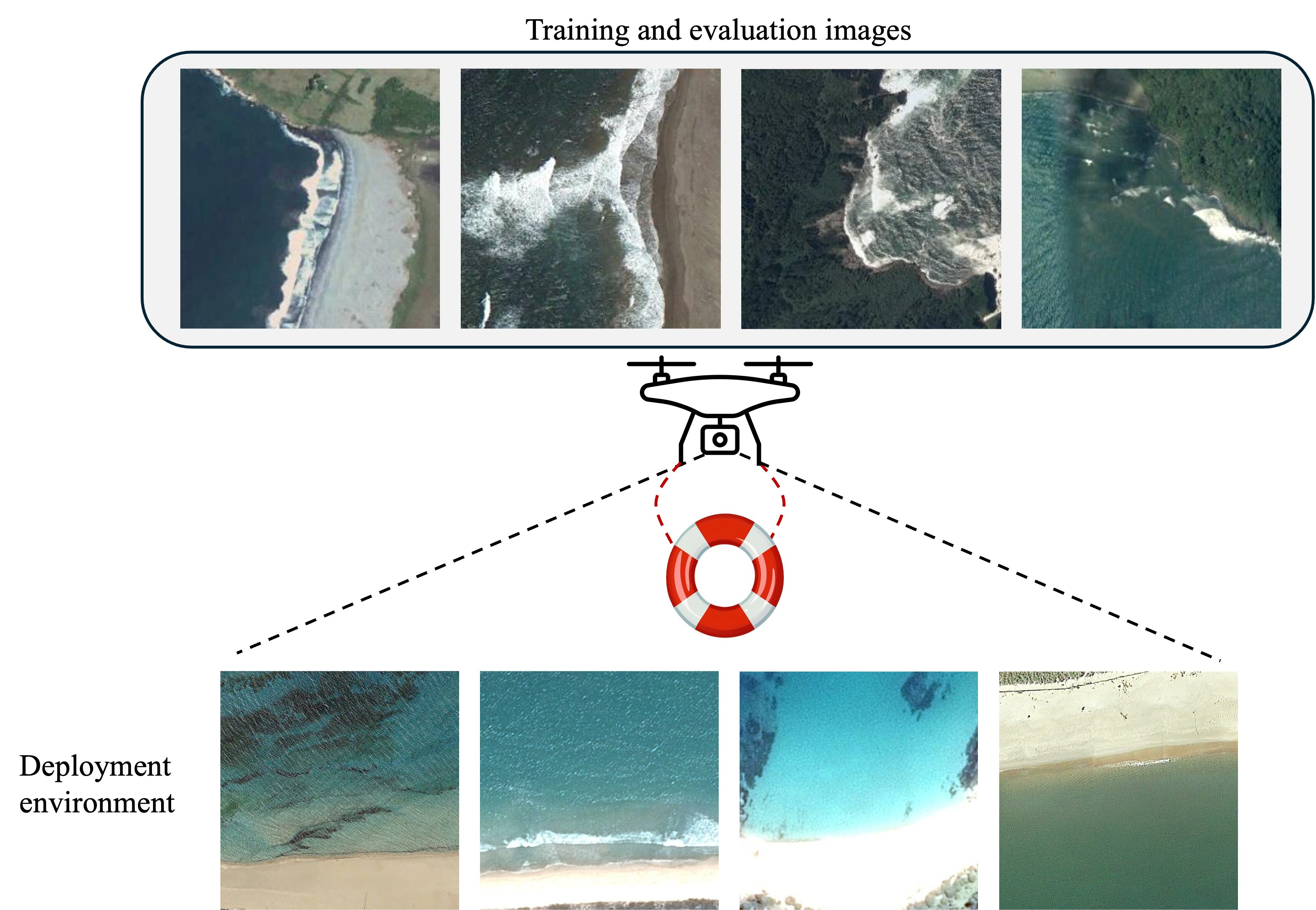}
   \caption{Illustration of potential distribution shift for a perception model deployed on a UAV in a search-and-rescue mission. The coastal-scene examples are drawn from two different benchmarks: training images are from RESISC45 \cite{ref18}, while AID \cite{ref19} is used to represent a shifted deployment distribution.}
   \label{fig1}
\end{figure}

These considerations are important to models deployed on Unmanned Aerial Vehicles (UAVs) which are used in a plethora of applications, including disaster response and search-and-rescue missions \cite{ref11, ref12}. UAVs are especially valuable in such use cases because they can be deployed rapidly and cover vast geographic regions. However, distribution shift is particularly challenging in such ML systems as the visual conditions encountered after deployment may differ from those represented during development, as demonstrated in \Cref{fig1}. Further, these systems’ onboard computational resources are often limited, which constrains both the models’ capabilities and any additional monitoring mechanisms performed during operation. Performance degradation in these systems can be consequential in safety-critical applications, where inaccurate prediction may affect downstream decisions. Despite substantial progress in the literature \cite{ref13, ref14, ref15}, recent benchmarks suggest that even domain-specific foundation models may exhibit inconsistent robustness under different forms of distribution shift \cite{ref16}, motivating further research on practical mechanisms for detecting distributional changes as they arise.

In this paper, we conduct a case study examining three open-source libraries whose detection capabilities can be readily used by practitioners: (i) Alibi-Detect \cite{ref6}; (ii) NannyML \cite{ref7}; and (iii) PyTorch-OOD \cite{ref8}. We acknowledge that other monitoring tools are available; however, a comprehensive inclusion of all existing tools is beyond the scope of this short paper. Instead, we use these three tools here as representative examples to study how practical detection approaches perform not only in terms of shift-detection, but also in terms of detector latency. Our study therefore aims to provide an empirical perspective on the trade-offs that arise when existing open-source detection tools and methods are considered for deployment in UAV-based and earth observation systems.
\section{Case-Study Overview}
\label{sec:case_study}

\begin{figure}[t]
  \centering
   \includegraphics[width=0.8\linewidth]{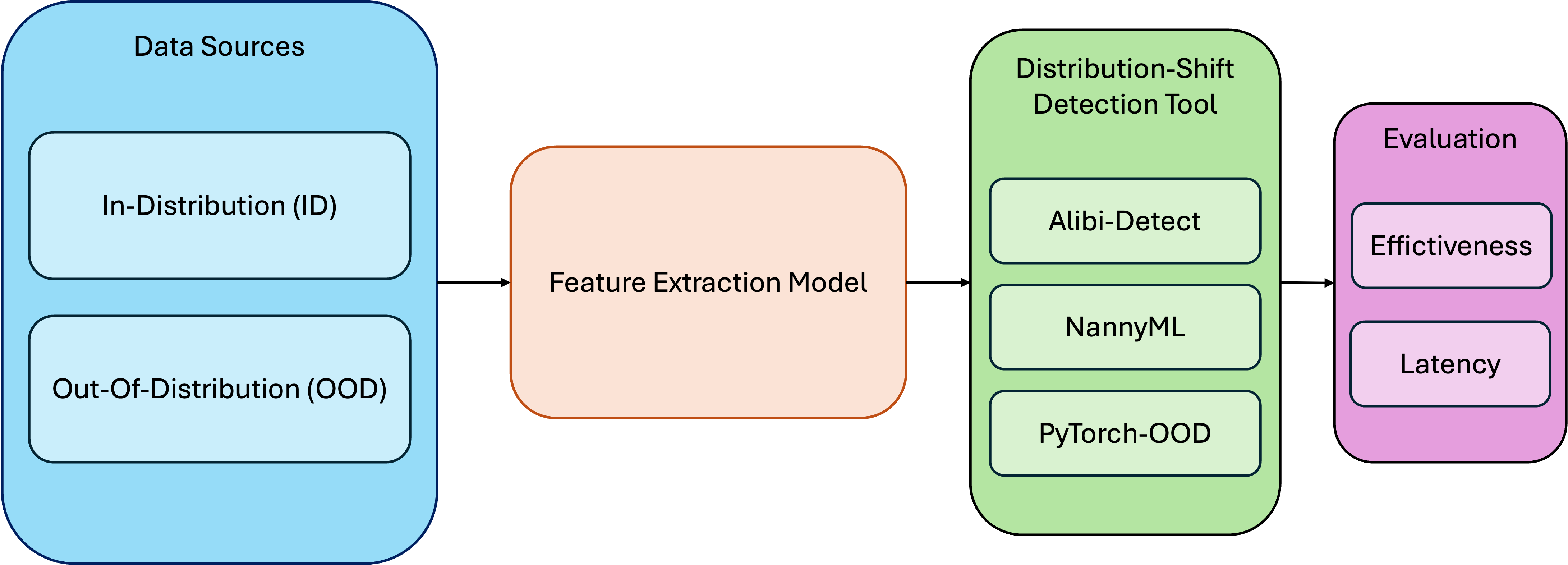}

   \caption{Overview of the proposed case-study design.}
   \label{fig2}
\end{figure}

\textbf{Analysis design.} Our case study is designed to examine distribution-shift detection from a post-deployment perspective, as illustrated in \Cref{fig2}. We first train a classification model using an in-distribution (ID) dataset, while data from separate sources are reserved to represent shifted distributions. After training, the model is used as a feature extractor to obtain representations from both data sources. These representations are then provided to the considered tools. The resulting detections are then assessed in terms of both their effectiveness in identifying distributional changes and the computational latency required to produce the corresponding detection decision. 

\section{Experimental Design}
\label{sec:exp}

\textbf{Datasets.} To examine distribution shift in a deployment-oriented setting, we use NWPU-RESISC45 (RESISC45) \cite{ref18} as the ID dataset and AID \cite{ref19} and UCMerced Land Use \cite{ref20} as the OOD datasets. These aerial scene-classification datasets were collected independently and differ in their geographic coverage and acquisition characteristics. Since their original label spaces are not identical, we restrict each cross-dataset experiment to classes with semantically equivalent labels. More details regarding the shared label spaces are provided in Appendix A.

\textbf{Methods and baselines.} Detecting distribution shifts can broadly be categorized as: (i) window-level (or batch-level) detection, where a statistical test compares a window of observations with reference dataset to determine whether their distributions differ; and (ii) sample-level detection, where individual observations are evaluated independently, commonly using anomaly or out-of-distribution (OOD) scores.  

From Alibi-Detect, we evaluate four drift detection methods available through its interface: Kolmogorov–Smirnov (KS), Cramér–von Mises (CvM), Least-Squares Density Difference (LSDD), and Maximum Mean Discrepancy (MMD). From NannyML, we evaluate four approaches: Hellinger Distance (Hellinger), Jensen–Shannon Distance (JSD), Kolmogorov–Smirnov (KS), and Wasserstein Distance (Wasserstein). Among the methods included in our analysis, interestingly, KS is the only statistical test common to both libraries, while the remaining methods differ. This experimental setup serves two purposes: (i) to examine how the same statistical test behaves across different library implementations, where implementation choices may affect both detection performance and latency; and (ii) to evaluate the overall practical behavior of each monitoring tool across the methods considered, rather than to perform a one-to-one comparison of identical statistical tests.

Finally, we examine a broader set of detection methods that operate on different model outputs or internal representations provided by PyTorch-OOD: (i) feature-based methods, including Mahalanobis, and NCI; (ii) logit-based methods, including MSP, and MaxLogits; and (iii) activation-based methods, including ReAct, and ASH.

\textbf{Implementation details.} We consider three model architectures: ResNet18, ResNet50, and ViT-B/16. All models are initialized with ImageNet-pretrained weights, with their classification heads replaced to match the corresponding label space. As input to the detection methods, we use representations extracted immediately before the classification layer. All three architectures are trained for 10 epochs with a batch size of 64, a learning rate of \(10^{-4}\), and weight decay of \(10^{-4}\). The experiments are conducted on an NVIDIA Tesla T4 GPU with 14 GB of GPU memory and 62 GB of system memory.

\textbf{Evaluation protocols.} For sample-level OOD detection, we evaluate each method using the area under the receiver operating characteristic curve (AUROC) and the false-positive rate at 95\% true-positive rate (FPR95). On the other hand, shifted samples are processed in non-overlapping windows of N $\in$ {16, 32, 64} for window-level methods. We report the shift-detection rate, defined as the proportion of shifted windows for which the detector identifies a statistically significant distributional change.

We additionally measure the computational latency of each detector in our analysis. For sample-level methods, latency is measured one observation at a time, whereas for window-level methods, latency corresponds to the detector computation required to process an already-accumulated observation in a window. GPU synchronization is performed before and after each timed operation, where applicable. We report detector-only latency, whereas model inference, feature extraction, data loading, and, for window-level methods the time required to accumulate N incoming samples, are excluded from our latency analysis.

\section{Results and Analysis}
\label{sec:results}

\textbf{Alibi-Detect vs. NannyML.} We first examine whether the cross-dataset shifts affect the predictive performance of the underlying classification models. As shown in \Cref{rs_accuracy}, all three architectures achieve high accuracy on the held-out RESISC45 ID data, but their performance decreases when evaluated on the target datasets. The degradation is more pronounced for UCMerced, whereas a smaller decrease is observed for AID. These results indicate that the cross-dataset shifts affect the predictive performance of all three architectures, although the magnitude of degradation varies across both target datasets and model architectures. This motivates examining whether the corresponding distributional changes can be reliably detected during deployment by the monitoring tools considered in our analysis and how the effectiveness of detection varies with their configurations.

\begin{table}[t]
\centering
\caption{Classification accuracy (\%) under cross-dataset remote-sensing shifts.}
\label{rs_accuracy}
\footnotesize
\setlength{\tabcolsep}{6.10pt}
\renewcommand{\arraystretch}{0.95}

\begin{tabular}{lcccc}
\toprule
& \multicolumn{2}{c}{\textbf{RESISC45 $\rightarrow$ AID}}
& \multicolumn{2}{c}{\textbf{RESISC45 $\rightarrow$ UCM}} \\
\cmidrule(lr){2-3}\cmidrule(lr){4-5}

\textbf{Model}
& \textbf{ID} & \textbf{Target}
& \textbf{ID} & \textbf{Target} \\
\midrule

ResNet-18 & 97.47 & 88.21 & 98.21 & 84.37 \\
ResNet-50 & 97.68 & 90.98 & 98.42 & 88.16 \\
ViT-B/16  & 97.89 & 93.34 & 98.84 & 91.47 \\

\bottomrule
\end{tabular}
\end{table}

\begin{figure}[t]
  \centering
   \includegraphics[width=0.8\linewidth]{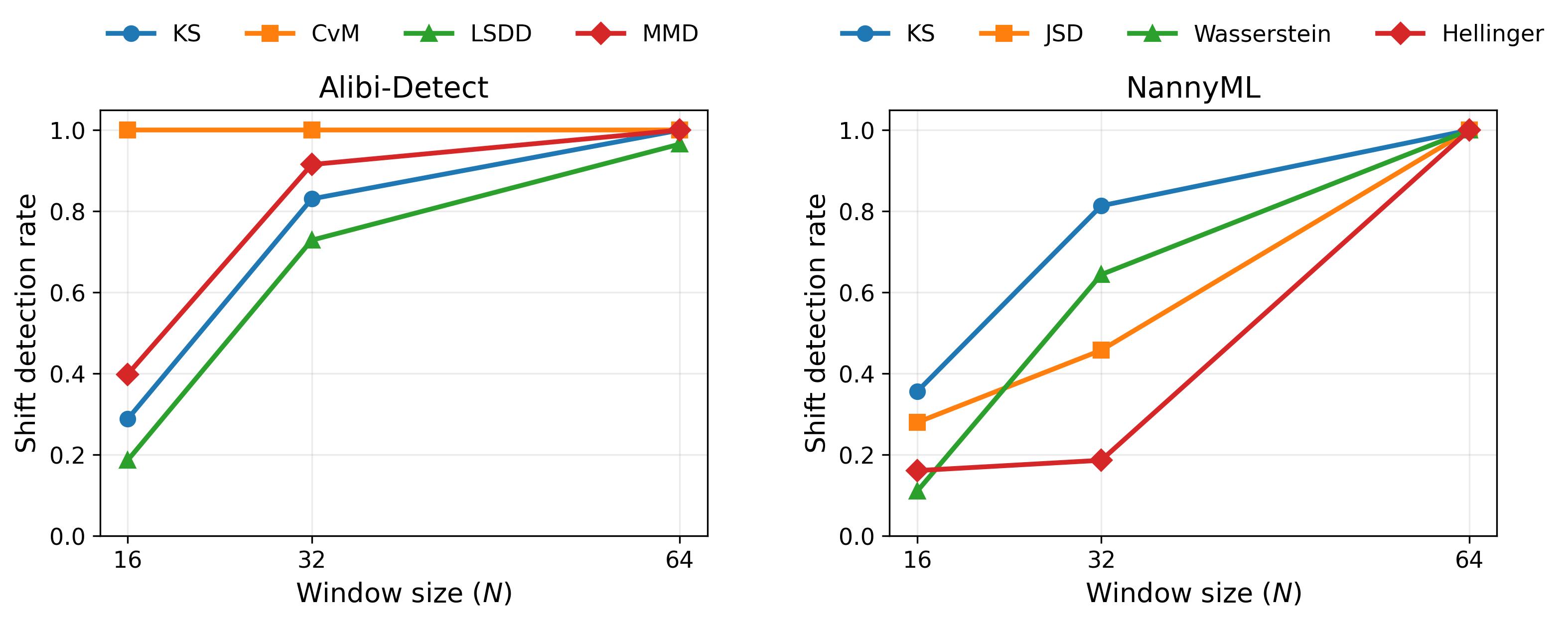}
   \caption{Window-level shift-detection rate for ResNet18 and RESISC45$\rightarrow$UCMerced shift setting.}
   \label{fig3}
\end{figure}

In \Cref{fig3}, we present the results for the RESISC45$\rightarrow$UCMerced shift setting using ResNet18 as the feature extractor. Across both Alibi-Detect and NannyML, the general trend is that the shift-detection rate improves as the window size increases. This behavior is expected because larger windows provide the statistical tests with more observations from the deployment distribution, making distributional differences easier to distinguish. However, the extent to which detection improves with window size varies substantially across models, methods and target datasets. For instance, CvM from Alibi-Detect achieves consistently high detection rates even at the smallest evaluated window size, whereas other methods require more observations to achieve reliable detection. This behavior is particularly evident for NannyML's Hellinger detector, whose detection rate is low at smaller windows and increases as the window size grows. Importantly, our additional experiments shown in Appendix B demonstrate that window size alone does not determine detection accuracy, as the behavior of the same detector can vary across models and target shifts. Therefore, methods should be selected based on their effectiveness across the deployment conditions relevant to the considered application.

\begin{figure}[t]
  \centering
   \includegraphics[width=0.8\linewidth]{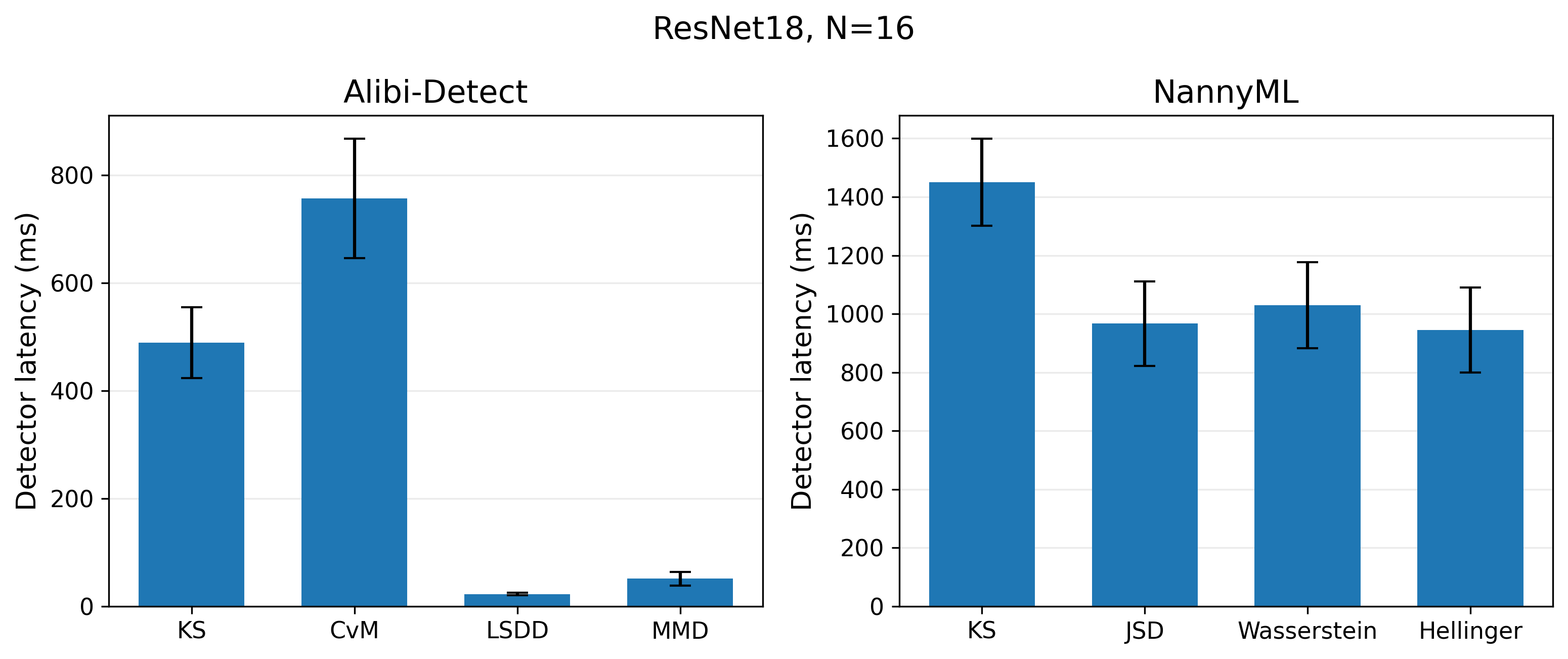}

   \caption{Detector latency for ResNet18 at window size N=16. Results of more models and larger windows are in Appendix C.}
   \label{fig4}
\end{figure}

We also analyze the computational cost of the considered methods to assess their practical viability during deployment. Although larger windows generally improve detection performance, they introduce additional delay in two ways: (i) sample-accumulation delay, corresponding to the time required to collect enough observations to form a complete window; and (ii) detector computation time, corresponding to the time required by the monitoring method itself to process the accumulated window. Both components contribute to the overall time-to-detection and may become important in time-sensitive applications. However, we exclude sample-accumulation delay from our measurements because it is application dependent and is influenced by factors such as the data-arrival rate and sensing configuration. Instead, our analysis focuses on evaluating the computational latency of each detector provided by the considered monitoring tools. This setup allows us to isolate the computational overhead of the detection methods from other sources of delay that may vary across deployment environments, hardware, and application-specific data-arrival rates. Nevertheless, these additional time components must be considered in practice when assessing whether a method can satisfy the deployment requirements.

Our results in \Cref{fig4} show that similar methods can exhibit different processing times across libraries because their implementations may differ. This is evident for KS, whose Alibi-Detect implementation is faster than the corresponding NannyML implementation under our setup. Further, the computational backend and hardware used by each implementation also appear to play a role in detector latency. The evaluated NannyML methods execute on the CPU, as do the KS and CvM implementations from Alibi-Detect, whereas the Alibi-Detect MMD and LSDD implementations use a GPU-enabled operations. The latter achieve lower processing times in our experiments, illustrating that implementation and hardware characteristics can be as important as the underlying statistical method. Taken together with the results in \Cref{fig3}, window-level monitoring methods can provide effective detection, but their practical suitability depends not only on detection accuracy but also on the time required to produce a decision. In time-sensitive UAV applications, high detector latency may limit the usefulness of the methods at hand, motivating further work toward practical monitoring approaches.

\textbf{PyTorch-OOD.} In contrast to methods provided by Alibi-Detect and NannyML, an alternative approach is to detect distribution shift at the level of individual observations. In this part of our analysis, we examine sample-level OOD detection methods provided by PyTorch-OOD. As shown in \Cref{tab:pytorch_ood_resnet18_ucm}, most of the evaluated methods operate at sub-millisecond or low-millisecond latency per sample. However, this computational advantage is accompanied by more modest performance under our considered experimental setup. The observed AUROC values indicate only moderate separation between ID and OOD samples, while FPR95 remains relatively high for many methods. 

\begin{table}[t]
\centering
\caption{Sample-level OOD detection performance using ResNet18 under the RESISC45$\rightarrow$UCMerced shift. Remaining experiments are in the Appendix.}
\label{tab:pytorch_ood_resnet18_ucm}

\setlength{\tabcolsep}{4pt}
\renewcommand{\arraystretch}{1.15}

\resizebox{\columnwidth}{!}{
\begin{tabular}{llccc}
\hline

\textbf{Category} &
\textbf{Method} &
\textbf{AUROC} $\uparrow$ &
\textbf{FPR95} $\downarrow$ &
\textbf{Latency (ms/sample)} $\downarrow$ \\

\hline

\multirow{2}{*}{Logit-based}
& MSP
& 0.724
& 0.830
& $0.248 \pm 0.051$ \\

& MaxLogit
& 0.726
& 0.834
& $0.146 \pm 0.029$ \\

\hline

\multirow{2}{*}{Feature-based}
& Mahalanobis
& 0.773
& 0.854
& $3.705 \pm 0.482$ \\

& NCI
& 0.726
& 0.880
& $0.674 \pm 0.148$ \\

\hline

\multirow{2}{*}{Activation-based}
& ReAct
& 0.732
& 0.839
& $0.687 \pm 0.127$ \\

& ASH
& 0.713
& 0.845
& $0.986 \pm 0.223$ \\

\hline
\end{tabular}
}
\end{table}

\section{Discussion}
\label{sec:discussion}

\textbf{Overall analysis.} Several existing tools provide accessible implementations of distribution-shift detection methods, allowing practitioners to select approaches that best suite the application at hand. NannyML primarily provides methods that execute on the CPU, which can limit their processing in time-sensitive deployments. Alibi-Detect provides similar capabilities, but some of its methods support GPU-enabled backends, which can reduce detector computation time. Generally, window-level methods from both libraries can provide effective detection once sufficient observations are accumulated, but their overall time-to-detection is affected by both sample-accumulation delay (i.e., data acquisition time) and computation time. Alternatively, PyTorch-OOD provides lightweight sample-level methods that can evaluate incoming observations immediately with low computational latency. However, our results indicate that this real-time processing may come at the cost of modest detection accuracy. These findings highlight the need for tools and methods that consider factors beyond detection accuracy, particularly in time-sensitive and resource-constrained systems such as real-time or near-real-time remote-sensing applications. For example, in high-speed navigation UAVs where end-to-end processing may need to occur within milliseconds \cite{ref22}, a slow or inaccurate detection component can reduce the value of the monitoring system if the decision arrives too late to support a subsequent response.

\textbf{Case-study limitations.} Our empirical analysis has several limitations that may bound the findings drawn from the reported results. First, the evaluated methods represent only a subset of the detection approaches available in the considered libraries. Experimenting with other methods may exhibit different characteristics. Second, our experiments consider only three open-source tools and other frameworks may differ in their implementations and characteristics. In our study, our objective was not to provide an exhaustive evaluation, but rather to highlight practical aspects that should be considered when evaluating monitoring tools. Third, although our experiments introduce naturally occurring distribution shifts, cross-dataset evaluation can combine multiple sources of variation and does not isolate a particular deployment shift. A more controlled remote-sensing benchmark would enable a more precise analysis of how different real-world shifts affect both predictive performance and the behavior of monitoring methods.
\section{Conclusion}
\label{sec:conclusion}

Our study examined a set of distribution-shift detection tools with emphasis on both detection effectiveness and latency. Using three open-source libraries, we evaluated both window-level and sample-level methods under deployment-oriented settings. Window-level methods generally provided stronger detection but incurred higher processing time, while sample-level methods offered faster responses with modest detection performance. These results highlight a practical trade-off between effectiveness and responsiveness. Although limited to three tools and an image-classification task, our study provides a conceptual basis for further research toward detection approaches that jointly consider reliability, efficiency, and response time, particularly for time-sensitive and safety-critical applications.
{
    \small
    \bibliographystyle{ieeenat_fullname}
    \bibliography{main}

@String(CVPRW= {IEEE Conf. Comput. Vis. Pattern Recog. Worksh.})

@String(CVPRW= {CVPRW})

@article{ref1,
  title={Dermatologist-level classification of skin cancer with deep neural networks},
  author={Esteva, Andre and Kuprel, Brett and Novoa, Roberto A and Ko, Justin and Swetter, Susan M and Blau, Helen M and Thrun, Sebastian},
  journal={nature},
  volume={542},
  number={7639},
  pages={115--118},
  year={2017},
  publisher={Nature Publishing Group UK London}
}

@article{ref2,
  title={Computer vision for autonomous vehicles: Problems, datasets and state of the art},
  author={Janai, Joel and G{\"u}ney, Fatma and Behl, Aseem and Geiger, Andreas},
  journal={Foundations and Trends in Computer Graphics and Vision},
  volume={12},
  number={1-3},
  pages={1--308},
  year={2020},
  publisher={Emerald Publishing Limited}
}

@article{ref3,
  title={Failing loudly: An empirical study of methods for detecting dataset shift},
  author={Rabanser, Stephan and G{\"u}nnemann, Stephan and Lipton, Zachary},
  journal={Advances in Neural Information Processing Systems},
  volume={32},
  year={2019}
}

@article{ref4,
  title={Openood v1. 5: Enhanced benchmark for out-of-distribution detection},
  author={Zhang, Jingyang and Yang, Jingkang and Wang, Pengyun and Wang, Haoqi and Lin, Yueqian and Zhang, Haoran and Sun, Yiyou and Du, Xuefeng and Li, Yixuan and Liu, Ziwei and others},
  journal={arXiv preprint arXiv:2306.09301},
  year={2023}
}

@article{ref5,
  title={Clip-powered domain generalization and domain adaptation: A comprehensive survey},
  author={Li, Jindong and Li, Yongguang and Fu, Yali and Liu, Jiahong and Liu, Yixin and Yang, Menglin and King, Irwin},
  journal={IEEE Transactions on Pattern Analysis and Machine Intelligence},
  year={2026},
  publisher={IEEE}
}

@software{ref6,
  title = {Alibi Detect: Algorithms for outlier, adversarial and drift detection},
  author = {Van Looveren, Arnaud and Klaise, Janis and Vacanti, Giovanni and Cobb, Oliver and Scillitoe, Ashley and Samoilescu, Robert and Athorne, Alex},
  url = {https://github.com/SeldonIO/alibi-detect},
  version = {0.13.0},
  date = {2025-12-11},
  year = {2019}
}

@misc{ref7,
    title = {{N}anny{ML} (release 0.13.1)},
    howpublished = {\url{https://github.com/NannyML/nannyml}},
    month = mar,
    year = 2023,
    note = {NannyML, Belgium, OHL.},
    key = {NannyML}
}

@inproceedings{ref8,
  title={Pytorch-ood: A library for out-of-distribution detection based on pytorch},
  author={Kirchheim, Konstantin and Filax, Marco and Ortmeier, Frank},
  booktitle={Proceedings of the IEEE/CVF Conference on Computer Vision and Pattern Recognition},
  pages={4351--4360},
  year={2022}
}

@inproceedings{ref9,
  title={Pyod 2: A python library for outlier detection with llm-powered model selection},
  author={Chen, Sihan and Qian, Zhuangzhuang and Siu, Wingchun and Hu, Xingcan and Li, Jiaqi and Li, Shawn and Qin, Yuehan and Yang, Tiankai and Xiao, Zhuo and Ye, Wanghao and others},
  booktitle={Companion Proceedings of the ACM on Web Conference 2025},
  pages={2807--2810},
  year={2025}
}

@article{ref99,
    author  = {Zhao, Yue and Nasrullah, Zain and Li, Zheng},
    title   = {PyOD: A Python Toolbox for Scalable Outlier Detection},
    journal = {Journal of Machine Learning Research},
    year    = {2019},
    volume  = {20},
    number  = {96},
    pages   = {1-7},
    url     = {http://jmlr.org/papers/v20/19-011.html}
}

@misc{ref10,
  author = {{Evidently AI}},
  title  = {{Evidently AI}},
  url    = {. https://www.evidentlyai.com/},
  note   = {Accessed: 2026-10-05}
}

@inproceedings{ref11,
  title={Detecting wildfires on UAVs with real-time segmentation trained by larger teacher models},
  author={Pesonen, Julius and Hakala, Teemu and Karjalainen, V{\"a}in{\"o} and Koivum{\"a}ki, Niko and Markelin, Lauri and Raita-Hakola, Anna-Maria and Suomalainen, Juha and P{\"o}l{\"o}nen, Ilkka and Honkavaara, Eija},
  booktitle={2025 IEEE/CVF Winter Conference on Applications of Computer Vision (WACV)},
  pages={5166--5176},
  year={2025},
  organization={IEEE}
}

@article{ref12,
  title={Unsupervised human detection with an embedded vision system on a fully autonomous UAV for search and rescue operations},
  author={Lygouras, Eleftherios and Santavas, Nicholas and Taitzoglou, Anastasios and Tarchanidis, Konstantinos and Mitropoulos, Athanasios and Gasteratos, Antonios},
  journal={Sensors},
  volume={19},
  number={16},
  pages={3542},
  year={2019},
  publisher={MDPI}
}

@article{ref13,
  title={An advanced Dirichlet prior network for out-of-distribution detection in remote sensing},
  author={Gawlikowski, Jakob and Saha, Sudipan and Kruspe, Anna and Zhu, Xiao Xiang},
  journal={IEEE Transactions on Geoscience and Remote Sensing},
  volume={60},
  pages={1--19},
  year={2022},
  publisher={IEEE}
}

@article{ref14,
  title={Evaluation of ten deep-learning-based out-of-distribution detection methods for remote sensing image scene classification},
  author={Li, Sicong and Li, Ning and Jing, Min and Ji, Chen and Cheng, Liang},
  journal={Remote Sensing},
  volume={16},
  number={9},
  pages={1501},
  year={2024},
  publisher={MDPI}
}

@inproceedings{ref15,
  title={Distribution shifts at scale: Out-of-distribution detection in earth observation},
  author={Ekim, Burak and Tadesse, Girmaw Abebe and Robinson, Caleb and Hacheme, Gilles and Schmitt, Michael and Dodhia, Rahul and Ferres, Juan M Lavista},
  booktitle={2025 IEEE/CVF Conference on Computer Vision and Pattern Recognition Workshops (CVPRW)},
  pages={2256--2265},
  year={2025},
  organization={IEEE}
}

@article{ref16,
  title={EarthShift: a benchmark for measuring robustness to real-world distribution shifts in Earth observation},
  author={Doerksen, Kelsey and Kerner, Hannah},
  journal={arXiv preprint arXiv:2605.29330},
  year={2026}
}

@article{ref18,
  title={Remote sensing image scene classification: Benchmark and state of the art},
  author={Cheng, Gong and Han, Junwei and Lu, Xiaoqiang},
  journal={Proceedings of the IEEE},
  volume={105},
  number={10},
  pages={1865--1883},
  year={2017},
  publisher={IEEE}
}

@article{ref19,
  title={AID: A benchmark data set for performance evaluation of aerial scene classification},
  author={Xia, Gui-Song and Hu, Jingwen and Hu, Fan and Shi, Baoguang and Bai, Xiang and Zhong, Yanfei and Zhang, Liangpei and Lu, Xiaoqiang},
  journal={IEEE transactions on geoscience and remote sensing},
  volume={55},
  number={7},
  pages={3965--3981},
  year={2017},
  publisher={IEEE}
}

@inproceedings{ref20,
  title={Bag-of-visual-words and spatial extensions for land-use classification},
  author={Yang, Yi and Newsam, Shawn},
  booktitle={Proceedings of the 18th SIGSPATIAL international conference on advances in geographic information systems},
  pages={270--279},
  year={2010}
}

@inproceedings{ref22,
  title={Towards low-latency event-based obstacle avoidance on a fpga-drone},
  author={Bonazzi, Pietro and Vogt, Christian and Jost, Michael and Khacef, Lyes and Paredes-Vall{\'e}s, Federico and Magno, Michele},
  booktitle={2025 IEEE/CVF Conference on Computer Vision and Pattern Recognition Workshops (CVPRW)},
  pages={4938--4946},
  year={2025},
  organization={IEEE}
}

@article{tonguz2025using,
  title={Using kolmogorov-smirnov distance for measuring distribution shift in machine learning},
  author={Tonguz, Ozan K and Taschin, Federico},
  journal={arXiv preprint arXiv:2510.15996},
  year={2025}
}

@article{taschin2025distribution,
  title={The Distribution Shift Problem in Transportation Networks Using Reinforcement Learning and AI},
  author={Taschin, Federico and Lazraq, Abderrahmane and Ozgunes, Inci and Tonguz, Ozan K},
  journal={IEEE Transactions on Intelligent Transportation Systems},
  year={2025},
  publisher={IEEE}
}

@article{tonguz2025measurement,
  title={On the measurement of distribution shift in machine learning systems},
  author={Tonguz, Ozan K and Taschin, Federico},
  journal={IEEE Intelligent Systems},
  volume={40},
  number={2},
  pages={45--54},
  year={2025},
  publisher={IEEE}
}
}

\end{document}